\documentclass[10pt,journal,compsoc]{IEEEtran}
\ifCLASSOPTIONcompsoc
  \usepackage[nocompress]{cite}
\else
  \usepackage{cite}
\fi
\usepackage{graphicx}
\usepackage{url}
\usepackage{amsfonts}
\usepackage{color}
\usepackage{booktabs}
\usepackage{subfig}
\newtheorem{example}{Example}

\begin{document}
\title{Theory-guided Data Science: A New Paradigm for Scientific Discovery from Data}
%
%
%
%

\author{Anuj Karpatne, Gowtham Atluri, James H. Faghmous, Michael Steinbach, Arindam Banerjee, \\  Auroop Ganguly, Shashi Shekhar, Nagiza Samatova, and Vipin Kumar
\IEEEcompsocitemizethanks{\IEEEcompsocthanksitem A. Karpatne is with the University of Minnesota.\protect\\
E-mail: karpa009@umn.edu
\IEEEcompsocthanksitem G. Atluri is with the University of Cincinnati.
\IEEEcompsocthanksitem J. H. Faghmous is with the Icahn School of Medicine at Mount Sinai.
\IEEEcompsocthanksitem M. Steinbach, A. Banerjee, S. Shekhar, and V. Kumar are with the University of Minnesota.
\IEEEcompsocthanksitem A. Ganguly is with the Northeastern University.
\IEEEcompsocthanksitem N. Samatova is with the North Carolina State University.
}
}

%
%

%

\IEEEtitleabstractindextext{%
\begin{abstract}

Data science models, although successful in a number of commercial domains, have had limited applicability in scientific problems involving complex physical phenomena. Theory-guided data science (TGDS) is an emerging paradigm that aims to leverage the wealth of scientific knowledge for improving the effectiveness of data science models in enabling scientific discovery. 
The overarching vision of TGDS is to introduce scientific consistency as an essential component for learning generalizable models. Further, by producing scientifically interpretable models, TGDS aims to advance our scientific understanding by discovering novel domain insights. Indeed, the paradigm of TGDS has started to gain prominence in a number of scientific disciplines such as turbulence modeling, material discovery, quantum chemistry, bio-medical science, bio-marker discovery, climate science, and hydrology. In this paper, we formally conceptualize the paradigm of TGDS and present a taxonomy of research themes in TGDS. We describe several approaches for integrating domain knowledge in different research themes using illustrative examples from different disciplines. We also highlight some of the promising avenues of novel research for realizing the full potential of theory-guided data science.

\end{abstract}

}

\maketitle

\IEEEdisplaynontitleabstractindextext

%
\IEEEpeerreviewmaketitle

\IEEEraisesectionheading{\section{Introduction}\label{sec:introduction}}

From satellites in space to wearable computing devices and from credit card transactions to electronic health-care records, the deluge of data \cite{bell2009beyond, economist2010data, james2011big} has pervaded every walk of life. Our ability to collect, store, and access large volumes of information is accelerating at unprecedented rates with better sensor technologies, more powerful computing platforms, and greater on-line connectivity. With the growing size of data, there has been a simultaneous revolution in the computational and statistical methods for processing and analyzing data, collectively referred to as the field of data science.
These advances have made long-lasting impacts on the way we sense, communicate, and make decisions \cite{halevy2009unreasonable}, a trend that is only expected to grow in the foreseeable future. 
Indeed, the start of twenty-first century may well be remembered in history as the ``golden age of data science.''

Apart from transforming commercial industries such as retail and advertising, data science is also beginning to play an important role in advancing scientific discovery. 
Historically, science has progressed by first generating hypotheses (or theories) and then collecting data to confirm or refute these hypotheses. However, in the big data era, ample data, which is being continuously collected without a specific theory or hypothesis in mind, offers further opportunity for discovering new knowledge.
Indeed, the role of data science in scientific disciplines is beginning to shift from providing simple analysis tools (e.g., detecting particles in Large Hadron Collider experiments \cite{roe2005boosted, castelvecchi2015artificial}) to providing full-fledged knowledge discovery frameworks (e.g., in bio-informatics \cite{baldi2001bioinformatics} and climate science \cite{faghmous2014big, faghmous2015computing}). Based on the success of data science in applications where Internet-scale data is available (with billions or even trillions of samples), e.g., natural language translation, optical character recognition, object tracking, and most recently, autonomous driving, there is a growing anticipation of similar accomplishments in scientific disciplines \cite{graham2008big, jonathan2011special, sejnowski2014putting}. To capture this excitement, some have even referred to the rise of data science in scientific disciplines as ``the end of theory''  \cite{Anderson2008}, the idea being that 
 the increasingly large amounts of data makes it possible to   build actionable models without using scientific theories. 

Unfortunately, this notion of black-box application of data science has met with limited success in scientific domains (e.g., \cite{caldwell2014statistical,Lazer2014, marcus2014eight}). A well-known example of the perils in using data science methods in a theory-agnostic manner is Google Flu Trends, where a data-driven model was learned to estimate the number of influenza-related physician visits based on the number of influenza-related Google search queries in the United States \cite{ginsberg2009detecting}. This model was built using search terms that were highly correlated with the flu propensity in the Center for Disease Control (CDC) data. Despite its initial success, this model later overestimated the flu propensity by more than a factor of two, as measured by the number of influenza-related doctor visits in subsequent years, according to CDC data \cite{Lazer2014}. 

There are two primary characteristics of knowledge discovery in scientific disciplines that have prevented data science models from reaching the level of success achieved in commercial domains. 
First, scientific problems are often under-constrained in nature as they suffer from paucity of representative training samples while involving a large number of physical variables.
Further, physical variables commonly show complex and non-stationary patterns that dynamically change over time. For this reason, the limited number of labeled instances available for training or cross-validation can often fail to represent the true nature of relationships in scientific problems. Hence, standard methods for assessing and ensuring generalizability of data science models may break down and lead to misleading conclusions. In particular, it is easy to learn spurious relationships that look deceptively good on training and test sets (even after using methods such as cross-validation), but do not generalize well outside the available labeled data.
This was one of the main reasons behind the failure of Google Flu Trends, since the data used for training the model in the first few years was not representative of the trends in subsequent years \cite{Lazer2014}.
The paucity of representative samples is one of the prime challenges that differentiates scientific problems from mainstream problems involving Internet-scale data such as language translation or object recognition, where large volumes of labeled or unlabeled data have been critical in the success of recent advancements in data science such as deep learning.


The second primary characteristic of scientific domains that have limited the success of black-box data science methods is the basic nature of scientific discovery. While a common end-goal of data science models is the generation of actionable models, the process of knowledge discovery in scientific domains does not end at that. 
Rather, it is the translation of learned patterns and relationships to \emph{interpretable} theories and hypotheses that leads to advancement of scientific knowledge, e.g., by explaining or discovering the physical cause-effect mechanisms between variables. Hence, even if a black-box model achieves somewhat more accurate performance but lacks the ability to deliver a mechanistic understanding of the underlying processes, it cannot be used as a basis for subsequent scientific developments. 
Further, an interpretable model, that is grounded by explainable theories, stands a better chance at safeguarding against the learning of spurious patterns from the data that lead to non-generalizable performance. This is especially important when dealing with problems that are critical in nature and associated with high risks (e.g., healthcare).

The limitations of black-box data science models in scientific disciplines motivate a novel paradigm that uses the unique capability of data science models to automatically learn patterns and models from large data, without ignoring the treasure of accumulated scientific knowledge.
We refer to this paradigm that attempts to integrate scientific knowledge and data science as \emph{theory-guided data science} (TGDS).
The paradigm of TGDS has already begun to show promise 
in scientific problems from diverse disciplines. Some examples include the discovery of novel climate patterns and relationships \cite{kawale2013graph, faghmous2015daily}, closure of knowledge gaps in turbulence modeling efforts \cite{singh2016machine, wang2016physics}, discovery of novel compounds in material science \cite{hautier2010finding, fischer2006predicting, curtarolo2013high},  design of density functionals in quantum chemistry \cite{li2015understanding}, improved imaging technologies in bio-medical science  \cite{wong2009active, xu2015robust}, discovery of genetic biomarkers \cite{liu2013accounting}, and the estimation of surface water dynamics at a global scale \cite{khandelwal2015post, rse}.
These efforts have been complemented with recent review papers 
\cite{faghmous2014big, wagner2016theory, MC.2014.335, npg-21-777-2014}, workshops (e.g., a 2016 conference on physics informed machine learning \cite{timl2016}) and industry initiatives (e.g., a recent IBM Research initiative on ``physical analytics'' \cite{ibm_pa}).

This paper attempts to build the foundations of theory-guided data science by presenting several ways of bringing scientific knowledge and data science models together, and illustrating them using examples of applications from diverse domains.
A major goal of this article is to formally conceptualize the paradigm of ``theory-guided data science'', where scientific theories are systematically integrated with data science models in the process of knowledge discovery.


The remainder of the article is structured as follows. Section \ref{sec:pgdm} provides an introduction to theory-guided data science and presents an overview of research themes in TGDS.
Sections \ref{sec:design}, \ref{sec:learning}, \ref{sec:refine}, \ref{sec:hybrid}, and \ref{sec:assimilation} describe several approaches in every research theme of TGDS, using illustrative examples from diverse disciplines.
Section \ref{sec:conclusion} provides concluding remarks.

\section{Theory-Guided Data Science} \label{sec:pgdm}
 

A common problem in scientific domains is to represent relationships among physical variables, e.g., the combustion pressure and launch velocity of a rocket or the shape of an aircraft wing and its resultant air drag.
The conventional approach for representing such relationships is to use models based on scientific knowledge, i.e., theory-based models, which encapsulate cause-effect relationships between variables that have either been empirically proven or theoretically deduced from first principles. These models can range from solving closed-form equations (e.g. using Navier--Stokes equation for studying laminar flow) to running computational simulations of dynamical systems (e.g. the use of numerical models in climate science, hydrology, and turbulence modeling). 
An alternate approach is to use a set of training examples involving input and output variables for learning a data science model that can automatically extract relationships between the variables.


\begin{figure}
\centering
\includegraphics[width=0.46\textwidth]{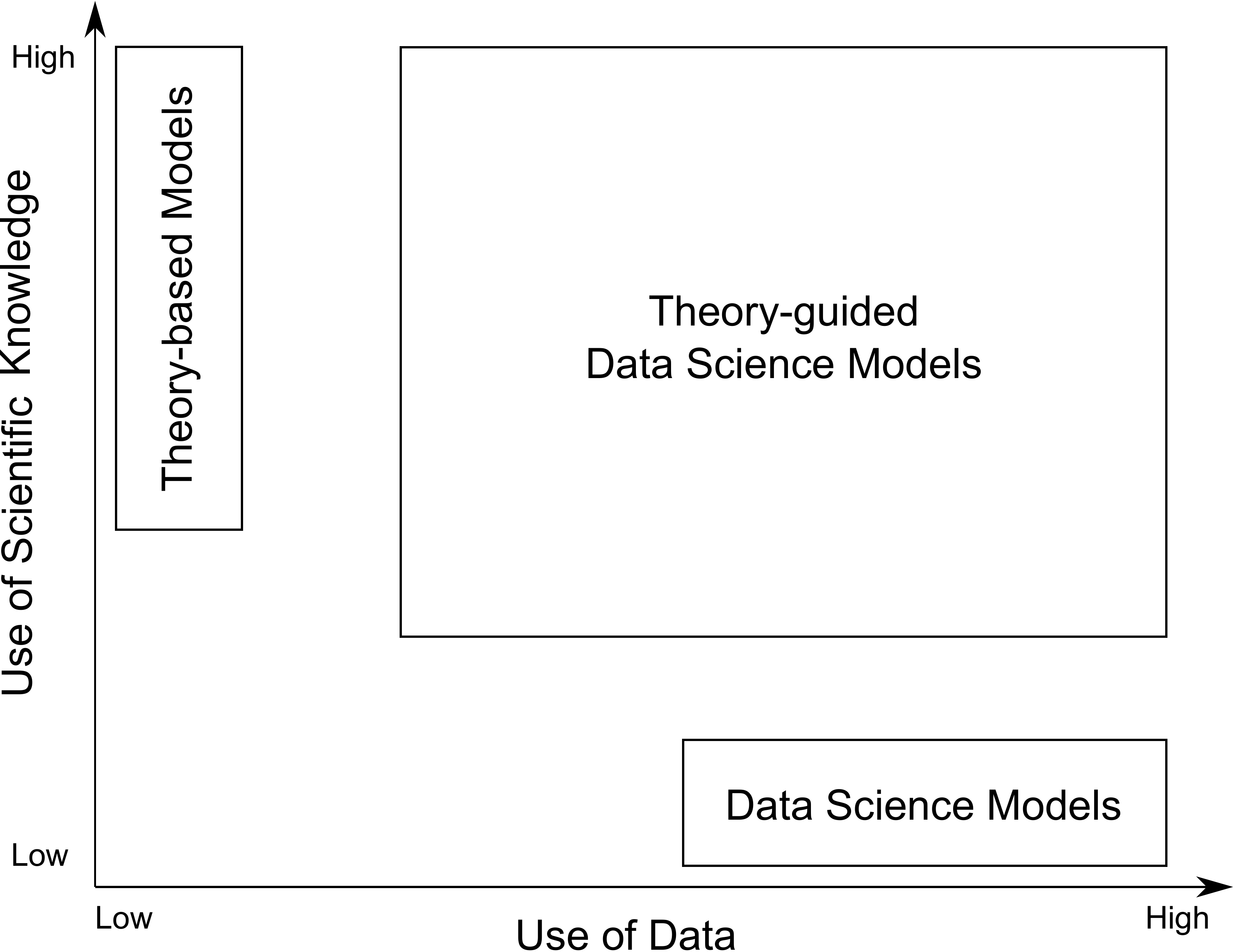}
\caption{A representation of knowledge discovery methods in scientific applications. The $x$-axis measures the use of data while the $y$-axis measures the use of scientific knowledge. Theory-guided data science explores the space of knowledge discovery that makes ample use of the available data while being observant of the underlying scientific knowledge.}
\label{fig:tgds}
\end{figure}


As depicted in Figure \ref{fig:tgds}, theory-based and data science models represent the two extremes of knowledge discovery, which depend on only one of the two sources of information available in any scientific problem, i.e., scientific knowledge or data. They both enjoy unique strengths and have found success in different types of applications. Theory-based models (see top-left corner of Figure \ref{fig:tgds}) are well-suited for representing processes that are conceptually well understood using known scientific principles.
On the other hand, traditional data science models mainly rely on the information contained in the data and thus reside in the bottom-right corner of Figure \ref{fig:tgds}. They have a wide range of applicability in domains where we have ample supply of representative data samples, e.g., in Internet-scale problems such as text mining and object recognition.

Despite their individual strengths, theory-based and data science models suffer from certain  deficiencies when applied in problems of great scientific relevance, where both theory and data are currently lacking.
For example, a number of scientific problems involve  processes that are not completely understood by our current body of knowledge, because of the inherent complexity of the processes. In such settings, theory-based models are often forced to make a number of simplifying assumptions about the physical processes, which not only leads to poor performance but also renders the model difficult to comprehend and analyze. We illustrate this scenario using the following example from hydrological modeling.

\begin{example}[Hydrological Modeling]

One of the primary objectives of hydrology is to study the processes responsible for the movement, distribution, and quality of water across the planet. Some examples of such processes include the discharge of water from the atmosphere via precipitation, and the infiltration of water underneath the Earth's surface, known as subsurface flow. 
Understanding subsurface flow is important as it is intricately linked with terrestrial ecosystem processes, agricultural water use, and sudden adverse events such as floods. 
However, our knowledge of subsurface flow using state-of-the-art hydrological models is quite limited \cite{ghasemizade2013subsurface}. This is mainly because subsurface flow operates in a regime that is difficult to measure directly using \emph{in-situ} sensors such as boreholes. In addition, subsurface flow involves a number of complex sub-processes that interact in non-linear ways, which are difficult to encapsulate in current theory-based models \cite{paniconi2015physically}.
Due to these challenges, existing hydrological models make use of a broad range of parameters in several weakly-informed physical equations. 
Thus, global hydrological models tend to show poor predictive performance in describing subsurface flow processes \cite{bierkens2015global}. In addition, they also lose physical interpretability due to the large number of model parameters that are difficult to interpret meaningfully with respect to the domain. 


\IEEEQED
\end{example}


If we apply ``black-box'' data science models  in scientific problems, we would notice a completely different set of issues arising due to the inadequacy of the available data in representing the complex spaces of hypotheses encountered in physical domains. Further, since most data science models can only capture associative relationships between variables, they do not fully serve the goal of understanding causative relationships in scientific problems.

Hence, neither a data-only nor a theory-only approach can be considered sufficient for knowledge discovery in complex scientific applications. Instead, there is a need to explore the continuum between theory-based and data science models, where both theory and data are used in a synergistic manner. 
The paradigm of \emph{theory-guided data science} (TGDS) attempts to 
address the 
shortcomings of data-only and theory-only models by seamlessly blending scientific knowledge in data science models (see Figure \ref{fig:tgds}). 
By integrating scientific knowledge in data science models, TGDS aims to learn dependencies that have a sufficient grounding in physical principles and thus have a better chance to represent causative relationships. 
TGDS further attempts to achieve better generalizability than models based purely on data by learning models that are {consistent with scientific principles}, termed as \emph{physically consistent models}.

\begin{figure}[ht]
\centering
\includegraphics[width=\linewidth]{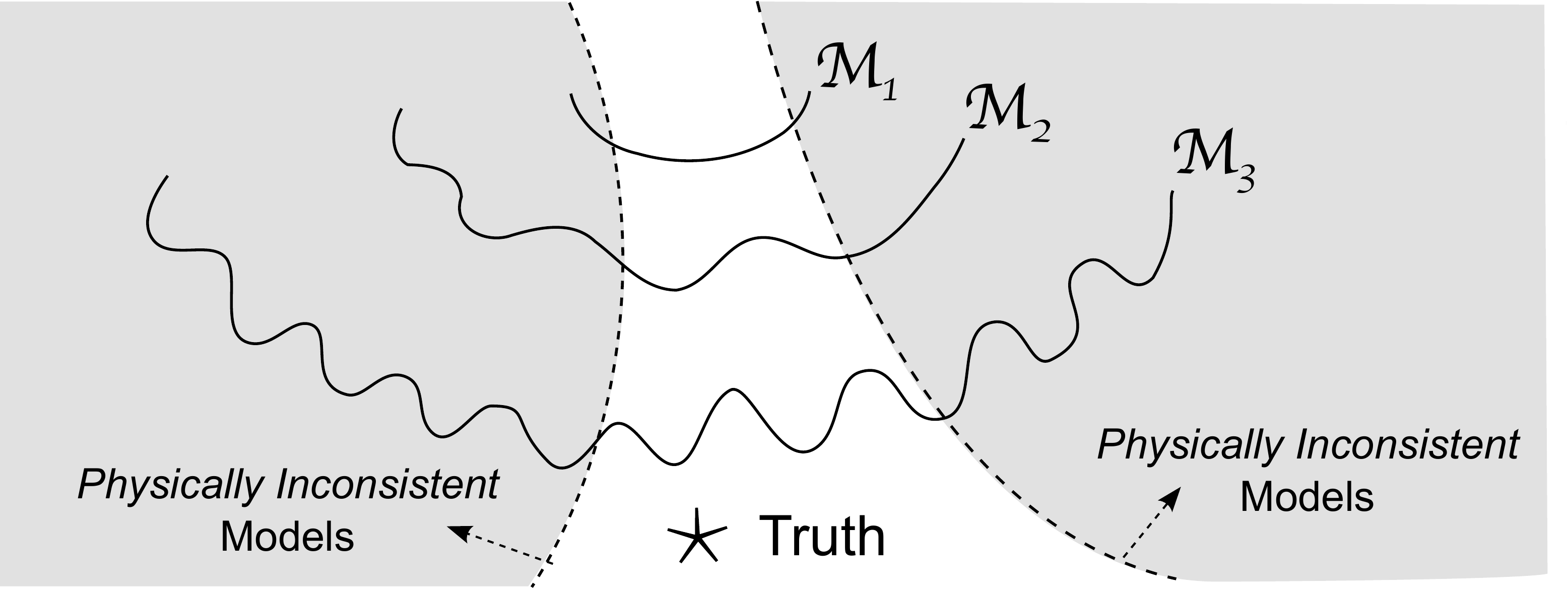}
\caption{Scientific knowledge can help in reducing the model variance by removing physically inconsistent solutions, without likely affecting their bias.}
\label{fig:constraints}
\end{figure}

To illustrate the role of ``consistency with scientific knowledge'' in ensuring better generalization performance, consider the example of learning a parametric model for a predictive learning problem using a limited supply of labeled samples. Ideally, we would like to learn a model that shows the best generalization performance over any unseen instance. Unfortunately, we can only observe the model performance on the available training set, which may not be truly representative of the true generalization performance (especially
when the training size is small). In recognition of this fact, a number of learning frameworks have been explored to favor the selection of \emph{simpler} models that may have lower accuracy on the training data (compared to more complex models) but are likely to have better generalization performance. This methodology, that builds on the well-known statistical principle of bias-variance trade-off \cite{friedman2001elements}, can be described using Figure \ref{fig:constraints}.

Figure \ref{fig:constraints} shows an abstract representation of a succession of model families  with varying levels of complexity (shown as curved lines), where $\mathcal{M}_1$ represents the set of least complex models while $\mathcal{M}_3$ contains highly complex models. Every point on the curved lines represents a model that a learning algorithm can arrive at, given a particular realization of training instances. The \emph{true} relationship between the input and output variables is depicted as a star in Figure \ref{fig:constraints}. 
We can observe that
the learned models belonging to $\mathcal{M}_3$, on average, are quite close to the true relationship.  However, even a small change in the training set can bring about large changes in the learned models of $\mathcal{M}_3$. 
Hence, $\mathcal{M}_3$ shows low \emph{bias} but high \emph{variance}. On the other hand, models belonging to $\mathcal{M}_1$ are quite robust to changes in the training set and thus show low variance. However, $\mathcal{M}_1$ shows high bias as its models are generally farther away from the true relationship as compared to models of $\mathcal{M}_3$. 
It is the trade-off between reducing bias and variance that is at the heart of a number of machine learning algorithms \cite{Tan2005, friedman2001elements, vapnik1998statistical}.


In scientific applications, there is another source of information that can be used to ensure the selection of generalizable models, which is the available scientific knowledge.
By pruning candidate models that are inconsistent with known scientific principles (shown as shaded regions in Figure \ref{fig:constraints}), we can significantly reduce the variance of models without likely affecting their bias. A learning algorithm can then be focused on the space of physically consistent models, leading to generalizable and scientifically interpretable models. Hence, one of the overarching visions of TGDS is to include physical \emph{consistency} as a critical component of model performance along with training accuracy and model complexity. 
This can be summarized in a simple way by the following revised objective of model performance in TGDS:
$$
\textrm{Performance} \propto \textrm{Accuracy} + \textrm{Simplicity} + {Consistency}.
$$

There are various ways of introducing physical consistency in data science models, in different forms and capacities. While some approaches attempt to naturally incorporate physical consistency in existing learning frameworks of data science models, others explore innovative ways of blending data science principles with theory-based models. In the following sections, we describe five broad categories of approaches for combining scientific knowledge with data science, that are illustrative of emerging examples of TGDS research in diverse disciplines. Note that many of these approaches can be applied together in multiple combinations for a particular problem, depending on the nature of scientific knowledge and the type of data science method. The five research themes of TGDS can be briefly summarized as follows.

First, scientific knowledge can be used in the design of model families to restrict the space of models to physically consistent solutions, e.g., in the selection of response and loss functions or in the design of model architectures. These techniques are discussed in Section \ref{sec:design}. Second, given a model family, we can also guide a learning algorithm to focus on physically consistent solutions. This can be achieved, for instance, by initializing the model with physically meaningful parameters, by encoding scientific knowledge as probabilistic relationships, by using domain-guided constraints, or with the help of regularization terms inspired by our physical understanding. These techniques are discussed in Section \ref{sec:learning}.
Third, the outputs of data science models can be refined using explicit or implicit scientific knowledge. This is discussed in Section \ref{sec:refine}. Fourth, another way of blending scientific knowledge and data science is to construct \emph{hybrid} models, where some aspects of the problem are modeled using theory-based components while other aspects are modeled using data science components. Techniques for constructing hybrid TGDS models are discussed in Section \ref{sec:hybrid}. Fifth, data science methods can also help in augmenting theory-based models to make effective use of observational data. These approaches are discussed in Section \ref{sec:assimilation}.

\section{Theory-guided Design of Data Science Models}
\label{sec:design}




An important decision in the learning of data science models is the choice of model family used for representing the relationships between input and response variables. 
In scientific applications, if the domain knowledge suggests a particular form of relationship between the inputs and outputs, care must be taken to ensure that the same form of relationship is used in the data science model.  Here, we discuss two different ways of using scientific knowledge in the design of data science models. First, we can use synergistic combinations of response and loss functions (e.g. in generalized linear models or artificial neural networks) that not only simplify the optimization process and thus lead to low training errors, but are also consistent with our physical understanding and hence result in generalizable solutions. Another way to infuse domain knowledge is by choosing a model architecture (e.g. the placement of layers in artificial neural networks) that is compliant with scientific knowledge. We discuss both these approaches in the following.

\subsection{Theory-guided Specification of Response}

Many data science models provide the option for specifying the form of relationship used for describing the response variable. For example, a generic family of models, which can represent a broad variety of relationships between input and response variables, is the generalized linear model (GLM). 
There are two basic building blocks in a GLM, the link function $g(.)$, and the probability distribution $P(y|\mathbf{x})$. Using these building blocks, the expected mean $\mu$ of the target variable $y$ is determined as a function of the weighted linear combination of inputs, $\mathbf{x}$, as follows:
\begin{eqnarray}
g(\mu) &=& \mathbf{w}^T \mathbf{x} + b, ~~\textrm{or equivalently,} \nonumber \\
\mu &=& g^{-1}(\mathbf{w}^T \mathbf{x} + b),
\end{eqnarray}
\begin{table}[b]
\centering
\caption{Table showing some commonly used combinations of link function and probability distribution functions in generalized linear models.}
\label{tab:glm}
\begin{tabular}{lcc}
\toprule
Name     & Link Function                      & Probability Distribution \\ \midrule
Linear   & $\mu$ & Gaussian                 \\
Poisson  & $\log(\mu)$                   & Poisson \\
Logistic & $\log(\mu/(1 - \mu))$                               & Binomial                       \\ \bottomrule
\end{tabular}
\end{table}
where $\mathbf{w}$ and $b$ and the parameters of GLM to be learned from the data.
Some common choices of link and probability distribution functions are listed in Table \ref{tab:glm}, resulting in varying types of regression models. 

To ensure the learning of GLMs that produce physically meaningful results, it is important to choose an appropriate specification of the response variable that matches with domain understanding. For example, while modeling response variables that show extreme effects (highly skewed distributions), e.g., occurrences of unusually severe floods and droughts, it would be inappropriate to assume the response variable to be Gaussian distributed (the standard assumption used in linear regression models). Instead, a regression model that uses the Gumbel distribution to model extreme values would be more accurate and physically meaningful. 
In general, the idea of specifying model response  using scientific principles can be explored in many types of learning algorithms. An example of theory-guided specification of response can be found in the field of ophthalmology, where the use of Zernike polynomials was explored by Twa et al. \cite{twa2005automated} for the classification of corneal shape using decision trees.

\subsection{Theory-guided Design of Model Architecture}
Scientific knowledge can also be used to influence the architecture of data science models. An example of a data science model that provides ample room for tuning the model architecture is artificial neural networks (ANN), which has recently gained widespread acceptance in several applications such as vision, speech, and language processing. 
There are a number of design considerations that influence the construction of an effective ANN model. Some examples include the number of hidden layers and the nature of connections among the layers, the sharing of model parameters among nodes, and the choice of activation and loss functions for effective model learning. 
Many of these design considerations are primarily motivated to simplify the learning procedure, minimize the training loss, and ensure robust generalization performance using statistical principles of regularization. 

There is a huge opportunity in informing these design considerations with our physical understanding of a problem, to obtain generalizable as well as scientifically interpretable results. For example, in an attempt to build a model of the brain that learns view-invariant features of human faces, the use of biologically plausible rules in ANN architectures was recently explored in \cite{cite-key}. It was observed that along with preserving view-invariance, such theory-guided ANN models were able to capture a known aspect of human neurology (namely, the mirror-symmetric tuning to head orientation) that was being missed by traditional ANN models. This made it possible to learn scientifically interpretable models of human cognition and thus advance our understanding of the inner workings of the brain. In the following, we describe two promising directions for using scientific knowledge while constructing ANN models: by using a modular design that is inspired by domain understanding, and by specifying the connections among the nodes in a physically consistent manner.

Domain knowledge can be used in the design of ANN models by decomposing the overall problem into modular sub-problems, each of which represents a different physical sub-process. Every sub-problem can then be learned using a different ANN model, whose inputs and outputs are connected with each other in accordance with the physical relationships among the sub-processes. For example, in order to describe the overall hydrological process of surface water discharge, we can learn modular ANN models for different sub-processes such as the atmospheric process of rainfall and evaporation, the process of surface water runoff, and the process related to groundwater seepage. Every ANN model can be fed with appropriately chosen domain features at the input and output layers. This will help in using the power of deep learning frameworks while following a high-level organization in the ANN architecture that is motivated by domain knowledge.


Domain knowledge can also be used in the design of ANN models by specifying node connections that  capture theory-guided dependencies among variables.
A number of variants of ANN have been explored to capture spatial and temporal dependencies between the input and output variables. For example, recurrent neural networks (RNN) are able to incorporate the sequential context of time in speech and language processing \cite{mikolov2010recurrent}. RNN models have been recently explored to capture notions of long and short term memory (LSTM) with the help of skip connections among nodes to model information delay \cite{sak2014long}. Such models can be used to incorporate time-varying domain characteristics in scientific applications. For example, while surface water runoff directly influences surface water discharge without any delay, groundwater runoff has a longer latency and contributes to the surface water discharge after some time lag. Such differences in time delay can be effectively modeled by a suitably designed LSTM model.
Another variant of ANN is the convolutional neural network (CNN) \cite{krizhevsky2012imagenet}, which has been widely applied in vision and image processing applications to capture spatial dependencies in the data. It further facilitates the sharing of model parameters so that the learned features are invariant to simple transformations such as scaling and transformation. Similar approaches can be explored to share the parameters (and thus reduce model complexity) over more generic similarity structures among the input features that are based on domain knowledge.

\section{Theory-guided Learning of Data Science Models}
\label{sec:learning}

Having chosen a suitable model design, the next step of model building involves navigating the search space of candidate models using a learning algorithm. 
In the following, we present four different ways of guiding the  learning algorithm to choose physically consistent models. First, we can use physically consistent solutions as initial points in iterative learning algorithms such as gradient descent methods. Second, we can restrict the space of probabilistic models with the help of theory-guided priors and relationships. Third, scientific knowledge can be used as constraints in optimization schemes for ensuring physical consistency. Fourth, scientific knowledge can be encoded as regularization terms in the objective function of learning algorithms. We describe each of these approaches in the following.

\subsection{Theory-guided Initialization}

Many learning algorithms that are iterative in nature require an initial choice of model parameters as a first step to commence the learning process. 
For such algorithms, an inferior initialization can lead to the learning of a poor model. Domain knowledge can help in the process of model initialization so that the learning algorithm is guided at an early stage to choose generalizable and physically consistent models.

An example of theory-guided initialization of model parameters includes a recent matrix completion approach for plant trait analysis \cite{schrodt2015bhpmf}, where the rows of the matrix correspond to plants from diverse environments while the columns correspond to plant traits such as leaf area, seed mass, and root length. Since observations about plant traits are sparsely available, such a plant trait matrix would be highly incomplete \cite{kattge2011try}. 
Filling the missing entries in a plant trait matrix can help us understand the characteristics of different plant species and their ability to adapt to varying environmental conditions.
A traditional data science approach to this problem is to use matrix completion algorithms that have found great success in online recommender systems \cite{melville2011recommender}. However, many of these algorithms are iterative in nature and use fixed or random values to initialize the matrix. In the presence of domain knowledge, we can improve these algorithms by using the species mean of every attribute as  initial values in the matrix completion process. This relies on the basic principle that the species mean provides a robust estimate of the average behavior across all organisms. This approach has been shown to provide significant improvements in the accuracy of predicting plant traits over traditional methods \cite{schrodt2015bhpmf}.
Changes from the species mean can also be learned using subsequent matrix completion operations, which could be physically interpreted as the effect of varying environmental conditions on plant traits.


One of the data science models that requires special efforts in choosing an appropriate combination of initial model parameters is the artificial neural network, which is known to be  susceptible to getting stuck at local minimas, saddle points, and flat regions in the loss curve.
In the era of deep learning, much progress has been made to avoid the problem of inferior ANN initialization with the help of \emph{pretraining} strategies. The basic idea of these strategies is to train the ANN model over a simpler problem (with ample availability of representative data) and use the trained model to initialize the learning for the original problem.
These pretraining strategies have made major impact on our ability to learn complex hierarchies of features in several application domains such as speech and image processing.
However, they rely on plentiful amounts of unlabeled or labeled data and hence are not directly applicable in scientific domains where the data sizes are small relative to the number of variables. One way to address this challenge is by devising novel pretraining strategies where computational simulations of theory-based models are used to initialize the ANN model. This can be especially useful when theory-based models can produce approximate simulations quickly, e.g., approximate model simulations of turbulent flow (see Example \ref{ex:turbulence}). 
Such pretrained theory-guided ANN models can then be fine-tuned using expert-quality ground truth.

\subsection{Theory-guided Probabilistic Models}

Probabilistic graphical models provide a natural way to encode domain-specific relationships among variables as edges between nodes representing the variables. 
However, manually encoding domain knowledge in graphical models requires a great deal of expert supervision, which can be cumbersome for problems involving a large number of variables with complex interactions--a common feature of scientific problems. In the presence of a large number of nodes, it is common to apply automated graph estimation techniques such as the use of graph Lasso \cite{friedman2008sparse}. The basic objective of such techniques is to estimate a sparse inverse covariance matrix that maximizes the model likelihood given the data. To assist such techniques with scientific knowledge, a promising research direction is to explore graph estimation techniques that maximize data likelihood while limiting the search to physically consistent solutions.




Another approach to reduce the variance of model parameters (and thus avoid model overfitting) is to introduce priors in the model space. 
An example of the use of theory-guided priors is the problem of non-invasive electrophysiological imaging of the heart. In this problem, the electrical activity within the walls of the heart needs to be predicted based on the ECG signal measured on the torso of a subject. There are approximately 2000 locations in the walls of the heart where electrical activity needs to be predicted, based on ECG data collected from approximately 100 electrodes on the torso. Given the large space of model parameters and the paucity of labeled examples with ground-truth information, a traditional black-box model that only uses the information contained in the data is highly prone to learning spurious patterns. However, apart from the knowledge contained in the data, we also have domain knowledge (represented using electrophysiological equations) about how electrical signals are transmitted within the heart via the myocardial fibre structure. These equations can be used to determine the spatial distribution of the electric signals in the heart at time $t$ based on the predicted electric signals at $t-1$. 
Incorporating such theory-guided spatial distributions as priors and using it along with externally collected ECG data in a hierarchical Bayesian model has been shown to provide promising results over traditional data science models \cite{wong2009active, xu2015robust}. Another example of theory-guided priors can be found in the field of geophysics \cite{denli2014multi}, where the knowledge of convection-diffusion equations was used as priors for determining the connectivity structure of subsurface aquifers.


\subsection{Theory-guided Constrained Optimization}

Constrained optimization techniques are extensively used in data science models for restricting the space of model parameters. For example, support vector machines use constraints for ensuring separability among the classes, while maximizing the margin of the hyperplane. 
There is also a rich literature on constraint-based pattern mining \cite{boulicaut2005constraint, pei2002constrained} and clustering \cite{basu2008constrained}. 
The use of constraints provides a natural way to integrate domain knowledge in the learning of data science models.
In scientific applications where theory-based constraints can be represented using linear equality or inequality conditions, they can be readily integrated in existing constrained optimization formulations, which are known to provide computationally efficient solutions especially when the objective function is convex.


 However, many scientific problems involve constraints that are represented in complex forms, e.g.,  using partial differential equations (PDE) or non-linear transformations of variables, which are not easily handled by traditional constrained optimization methods.
 For example, the Naiver--stokes equation for momentum expresses the following constraint between the flow velocity $\mathbf{v}$ and the fluid pressure ${p}$:
$$
\rho\Big(\frac{\partial\mathbf{u}}{\partial t} + \mathbf{u}\cdot \nabla \mathbf{u}\Big) = -\nabla p + \nabla \cdot (\mu (\nabla \mathbf {u}  + (\nabla \mathbf{u})^T) - \frac{2}{3}\mu (\nabla \cdot \mathbf{u})\mathbf{I}),
$$
where $\rho$ is the fluid density, $\mu$ is the fluid dynamic viscosity, and $\nabla$ represents the gradient operator with respect to the spatial coordinates.

To utilize such complex forms of constraints in data science models, it is necessary to develop constrained optimization techniques that can use common forms of partial differential equations encountered in scientific disciplines. An example of a data-driven approach that uses domain-driven PDEs can be found in a recent work in climate science \cite{majda2012physics, majda2012fundamental}, where physically constrained time-series regression models were developed to incorporate memory effects in time as well as the nonlinear noise arising from energy-conserving interactions.

In the following, we present detailed discussions of two illustrative examples of the use of theory-guided constraints. While Example \ref{ex:chemistry} explores the use of constraints for predicting electron density in computational chemistry, Example \ref{ex:water1} explores the use of elevation-based constraints among locations for mapping surface water dynamics.


\begin{example}[Computational Chemistry]
\label{ex:chemistry}

In computational chemistry, solving Schr\"{o}dinger's equation is at the basis of all quantum mechanical calculations for predicting the properties of solids and molecules. Schr\"{o}dinger's equation can be expressed as
\begin{eqnarray}
\mathbf{H}\Psi &=& \mathbf{E}\Psi, \\
&=& (\mathbf{T} + \mathbf{U} + \mathbf{V}) \Psi,
\end{eqnarray}
where $\mathbf{H}$ is the electronic Hamiltonian operator, $\Psi$ is the wavefunction that describes the quantum state of the system, and $\mathbf{E}$ is the total energy consisting of three terms, the kinetic energy, $\mathbf{T}$, the electron-electron interaction energy, $\mathbf{U}$, and the potential energy arising due to external fields, $\mathbf{V}$ (e.g., due to positively charged nuclei). Since the computational complexity in directly solving the Schr\"{o}dinger's equation grows rapidly with the number of particles, $N$,  it is infeasible for solving large many-particle systems in practical applications. 

To address this, a new class of quantum chemical modeling approaches was developed by Hohenberg and Kohn in 1964 \cite{hohenberg1964inhomogeneous}, which uses the electron density $n(r)$ as a basic primitive in all calculations, instead of the wavefunction $\Psi$. 
This has resulted in the rise of density functional theory (DFT) methods, which have become a standard tool for solving many-particle systems.
In DFT, every variable can be expressed as a functional of the electron density function $n(r)$ (where a functional is a function of functions). For example, the total energy $\mathbf{E}$ can be expressed in terms of functionals of $n(r)$ as follows:
\begin{equation}
\mathbf{E}[n] = \mathbf{T}[n] + \mathbf{U}[n] + \mathbf{V}[n].
\end{equation}
The density, $n_0(r)$, that leads to the lowest total energy, $\mathbf{E}[n_0]$, is known as the ground-state density of the system, which is a critical quantity to determine. 

However, obtaining $n_0(r)$ is challenging because of the interaction functional, $\mathbf{U}[n]$, whose exact form is unknown. Different approximations of the interaction term have been developed to solve for the ground-state density of a system, the most notable being the class of Kohn-Sham (KS) DFT methods. However, their performance is sensitive to the quality of approximation used in modeling the interactions. Also, KS DFT methods have a computational complexity of $O(N^3)$, which makes them challenging to apply on large systems.

To overcome the challenges in existing DFT methods, a recent work by Li et al. \cite{li2015understanding} explored the use of data science models to approximate $\mathbf{T}[n]$, and use such approximations to predict the ground-state density, $n_0(r)$. In this work, kernel ridge regression methods were used to model the kinetic energy, ${\mathbf{T}}[n]$, of a 4-particle system as a functional of its electron density, $n(r)$.
Having learned $\hat{\mathbf{T}}[n]$, we can obtain the ground-state energy, $n_0(r)$, using the following Euler-Lagrangian equation:
\begin{equation}
\frac{\delta \hat{\mathbf{T}}[n_0]}{\delta n_0(r)} = \mu - v(r), \label{eqn:euler}
\end{equation}
where $v(r)$ is the external potential and $\mu$ is an adjustable constant. 
This imposes a theory-guided constraint on the model learning, such that $\hat{\mathbf{T}}[n]$ must not only show good performance in predicting the kinetic energy, but should also accurately estimate the ground-state density, $n_0(r)$, using Equation \ref{eqn:euler}. A functional that adheres to this constraint can be called ``self-consistent.'' 

It was shown in \cite{li2015understanding} that a regression model that only focuses on minimizing the training error leads to highly inconsistent solutions of the ground-state density, and is thus not useful for quantum chemical calculations. This inconsistency can be traced to the inability of regression models in capturing functional derivative forms that are used in Equation \ref{eqn:euler}. In particular, the derivative of $\hat{\mathbf{T}}[n]$ can easily leave the space of densities observed in the training set, and thus arrive at ill-conditioned solutions especially when the training size is small. 

To overcome this limitation, a modified Euler-Lagrange constraint was proposed in \cite{li2015understanding}, which restricted the space of $n_0(r)$ to the density manifold observed in the training set. This helped in learning accurate as well as self-consistent ground-state densities using the knowledge contained in the data as well as domain theories. 

\IEEEQED
\end{example}




\begin{example}[Mapping Surface Water Dynamics]
\label{ex:water1}

Remote sensing data from Earth observing satellites presents a promising opportunity for monitoring the dynamics of surface water body extent at regular intervals of time. It is possible to build predictive models that use multi-spectral data from satellite images as input features to classify pixels of the image as water or land.
However, these models are challenged by the poor quality of labeled data, noise and missing values in remote sensing signals, and the inherent variability of water and land classes over space and time \cite{karpatne2016global, karpatne2016monitoring}. 

To address these challenges, there is an opportunity for improving the quality of classification maps by using the domain knowledge that
water bodies have a concave elevation structure. Hence, locations at a lower elevation are filled up first before the water level reaches locations at higher elevations. Thus, if we have access to elevation information (e.g. from bathymetric measurements obtained via sonar instruments), we can use it to constrain the classifier so that it not only minimizes the training error in the feature space but also produces labels that are consistent with the elevation structure. To illustrate this, consider an example of a two-dimensional training set shown in Figure \ref{fig:depth1features}, where the squares and circles represent training instances belonging to water and land classes, respectively. Along with the features, we also have information about the elevation of every instance, shown using the intensity of colored points in Figure \ref{fig:depth1features}. 

\begin{figure}[h]
\centering
\subfloat[Distribution of water and land training samples from a specific water body in feature space. Shading reflects elevation information at the locations of training samples.]{
\includegraphics[width=\linewidth]{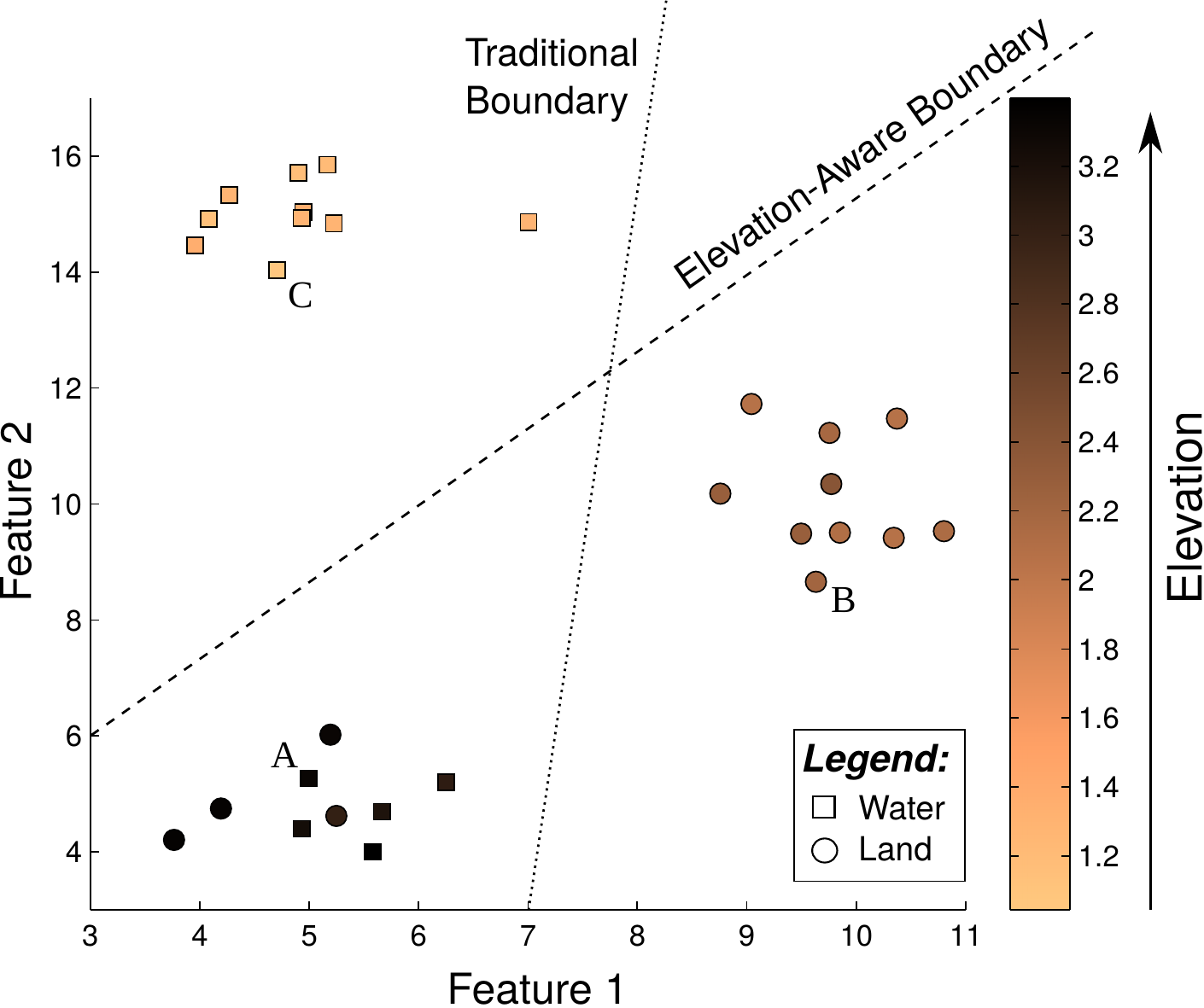}
\label{fig:depth1features}}
\\
\subfloat[Lake cross-section.]{
\includegraphics[width=0.7\linewidth]{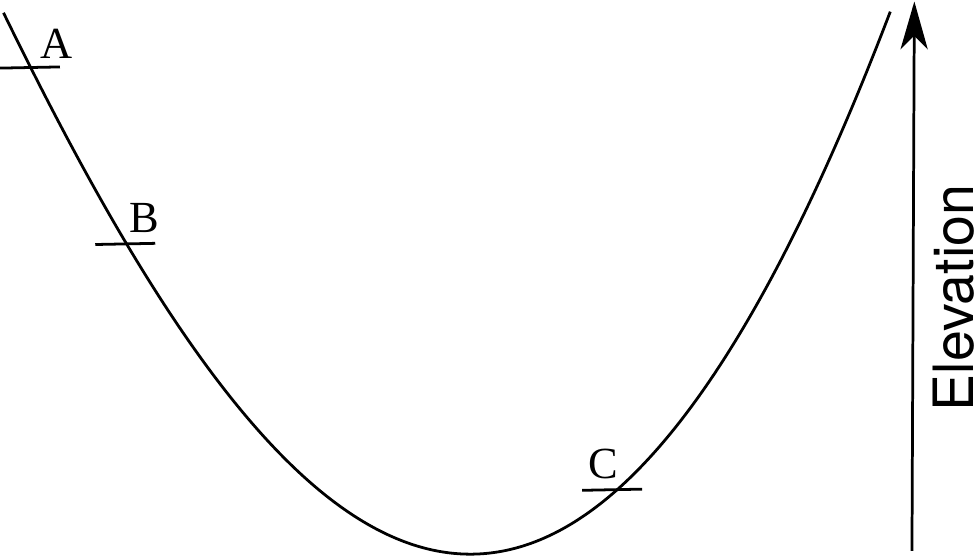}
\label{fig:depth1elev}}
\caption{An illustrative example of the use of elevation-based ordering (domain theory) for learning physically consistent classification boundaries of water and land. Along with the distribution of training instances in the feature space, we also have information about their elevation, as shown in  Figure \ref{fig:depth1features}). This information can be used to learn an elevation-aware classification boundary that produces physically viable labels, e.g. if $B$ is labeled as land, then $A$ must necessarily be labeled as land as it is at a higher elevation, as shown in Figure \ref{fig:depth1elev}.}
\label{fig:depth1}
\end{figure}

If we disregard the elevation information and learn a linear classifier to simply minimize the training errors, we would learn the decision boundary shown using a dotted line in Figure \ref{fig:depth1features}. This classifier would make some mistakes in the lower-left corner of the feature space, where the class confusion is difficult to resolve using a linear separator. However, if we use the elevation information, we can see that the entire group of instances in the lower lower-left corner has a higher elevation than the instances shown on the right (labeled as land), and are thus less likely to be filled with water.
For example, notice that location $A$ is at a higher elevation than both $B$ and $C$ (see Figure \ref{fig:depth1elev}). Hence, if $B$ is labeled as land, it would be inconsistent 
to classify $A$ as water and instead it should be classified as land. The use of such constraints can help in learning a generalizable classification model even with poorly labeled training data.


\IEEEQED
\end{example}

\subsection{Theory-guided Regularization}

One way to constrain the search space of model parameters is to use regularization terms in the objective function, which penalize the learning of overly complex models. A number of regularization techniques have been explored in the data science community to enforce different measures of model complexity. For example, minimizing the $L_p$ norm of model parameters has been extensively used for obtaining various effects of regularization in parametric model learning. While the $L_2$ norm has been used to avoid overly large parameter values in ridge regression and support vector machines, minimizing the $L_1$ norm results in the Lasso formulation and the Dantzig selector, both of which encode sparsity in the model parameters. 

However, these techniques are agnostic to the physical feasibility of the learned model and thus can lead to physically inconsistent solutions. For example, while predicting the elastic modulus using bond energy and melting point, Lasso may favor melting point over bond energy even though a direct causal link exists between bond energy and the modulus \cite{wagner2016theory}. This can result in the elimination of meaningful attributes and the selection of secondary attributes that are not directly relevant. 
Hence, there is a need to devise regularization techniques that can incorporate scientific knowledge to restrict the search space of model parameters. For example, instead of using the $L_p$ norm for regularization, we can find solutions on physically consistent sub-spaces of models. The Gaussian widths of such sub-spaces can be used as a regularization term in techniques such as the generalized Dantzig selector \cite{james2009generalized, chatterjee2014generalized}.
In the following, we describe two research directions for theory-guided regularization that have been explored in different applications: using variants of Lasso to incorporate domain-specific structure among parameters, and the use of multi-task learning formulations to account for the heterogeneity in data sub-populations.

The group Lasso \cite{yuan2006model} is a useful variant of Lasso that has been explored in problems involving structured attributes. It assumes the knowledge of a grouping structure among the attributes, where only a small number of groups are considered relevant. 
As an example in bio-marker discovery, the groups of attributes may correspond to sets of bio-markers that are related via a common biological pathway. 
Group Lasso helps in selecting physically meaningful groups of attributes in the data science models, and various extensions of group Lasso have been explored for handling different types of domain characteristics, e.g., overlapping group Lasso \cite{jacob2009group}, tree-guided group Lasso \cite{kim2010tree}, and sparse group Lasso \cite{friedman2010note}.

In recent work \cite{chatterjee2012sparse}, applications of sparse group Lasso were explored to model the domain characteristics of climate variables. In this work, climate variables observed over a range of spatial locations were used to predict a climate phenomenon of interest. By treating the set of variables observed at every location as a group, the use of group Lasso ensured that if a location is selected, all of the climate variables observed at that location will be used as relevant features. Such features thus represent meaningful (spatially coherent) regions in space that can be studied to identify physical pathways of relationships in climate science.

%
%




Another example of Lasso-based regularization that encodes domain knowledge can be found in the problem of discovering genetic markers for diseases. In this problem, data-driven approaches such as elastic nets are traditionally used to determine the relative importance of genetic markers in the context of a disease. However, geneticists understand that the relevant markers typically are located in close proximity on the genome sequence due to a property called linkage disequilibrium, which suggests that genetic information that is closely located travels together between generations of the population. This domain knowledge can be incorporated as a regularizer to ensure that the discovered genetic markers are typically located in close proximity on the genome. In fact, Liu and colleagues \cite{liu2013accounting} introduced a smoothed minimax concave penalty to Lasso that captured squared differences in regression coefficients between adjacent markers to ensure that the difference in genetic effects between adjacent markers is small.

Domain knowledge can also be used to guide the regularization of a multi-task learning (MTL) model, as explored for the problem of forest cover estimation in \cite{karpatne2014predictive}. In the presence of heterogeneity in data sub-populations, different groups of instances in the data show different relationships between the inputs and outputs. 
For example, different types of vegetation (e.g. forests, farms, and shrublands) may show varying responses to a target variable in remote sensing signals.
MTL provides a promising solution to handle sub-population heterogeneity in such cases, by treating the learning at every sub-population as a different task. Further, by sharing the learning at related tasks, MTL enforces a robust regularization on the learning across all tasks, even in the scarcity of training data.

However, most MTL formulations require explicit knowledge of the composition of every task and the similarity structure among the tasks, which is not always known in practical applications. For example, the exact number and distribution of vegetation types is often unavailable, and when they are known, they are available at varying granularties \cite{karpatne2016monitoring}. In recent work \cite{karpatne2014predictive}, the presence of heterogeneity due to varying vegetation types was first inferred by clustering vegetation time series, which was then used to induce similarity in the model parameters at related vegetation types. This resulted in an MTL formulation where the task structure was inferred using contextual variables, obtained using domain knowledge. 

\section{Theory-guided Refinement of Data Science Outputs}
\label{sec:refine}

Domain knowledge can also be used to refine the outputs of data science models so that they are in compliance with our current understanding of physical phenomena. This style of TGDS leverages scientific knowledge at the final stage of model building where the outputs of any data science model are made consistent with domain knowledge. In the following, we describe some of the approaches for refining data science outputs using domain knowledge that is either explicitly known (e.g. in the form of closed-form equations or model simulations) or implicitly available (e.g. in the form of latent constraints).

\subsection{Using Explicit Domain Knowledge}
Data science outputs are often refined to reduce the effect of noise and missing values and thus improve the overall quality of the results. For example, in the analysis of spatio-temporal data, there is a vast body of literature on refining model outputs to enforce spatial coherence and temporal smoothness among predictions. Data science outputs can  also be refined to improve a quality measure, e.g., in the discovery of frequent itemsets by \emph{pruning} candidate patterns. 
Building on these methods, a promising direction is to develop model refinement approaches that make ample use of domain knowledge, encoded in the form of scientific theories, for producing physically consistent results.



An example of theory-guided refinement of data science outputs can be found in the problem of material discovery, where the objective is to find novel materials and crystal structures that show a desirable property, e.g., their ability to filter gases or to serve as a catalyst. 
Traditional approaches for predicting crystal structure and properties rely on \emph{ab initio} calculations such as density functional theory methods. However, since the space of all possible materials is extremely large, it is impractical to perform computationally expensive \emph{ab initio} calculations on every material to estimate their structure and properties. 
Recently, a number of teams in material science have explored the use of probabilistic graphical models for predicting the structure and properties of a material, given a training database of materials with known structure and properties
\cite{hautier2010finding, fischer2006predicting, curtarolo2013high}. This provided a computationally efficient approach to reduce the space of candidate materials that show a desirable property, using the knowledge contained in the training data. The results of the data science models were then cross-checked using expensive \emph{ab initio} calculations to further refine the model outputs. This line of research has resulted in the discovery of a hundred new ternary oxide compounds that were previously unknown using traditional approaches \cite{hautier2010finding}, highlighting the effectiveness of TGDS in advancing scientific knowledge. 



\subsection{Using Implicit Domain Knowledge}

In scientific applications, the domain structure among the output variables may not always be known in the form of explicit equations that can be easily integrated in existing model refinement frameworks. This requires jointly solving the dual problem of inferring the domain constraints and using the learned constraints to refine model outputs.
We illustrate this using an example in mapping surface water dynamics, where implicit constraints among locations (based on a hidden elevation ordering) are estimated and leveraged for refining classification maps of water bodies.

\begin{example}[Post-processing using elevation constraints]
\label{ex:water0}

As described in Example \ref{ex:water1}, it is difficult to map the dynamics of surface water bodies by solely using the knowledge contained in remote sensing data, and there is promise in using information about the elevation structure of water bodies to assist classification models.
However, such information is seldom available at the desired granularity for most water bodies around the world. Hence, there is a need to infer the latent ordering among the locations (based on their elevation) so that they can be used to produce accurate and physically consistent labels. One way to achieve this is by using the history of imperfect water/land labels produced by a data science model at every location over a long period of time. In particular, a location that has been classified as water for a longer number of time-steps has a higher likelihood of being at a deeper location than a location that has been classified as water less frequently. 
This implicit elevation ordering, if extracted effectively, can help in improving the classification maps by post-processing the outputs to be consistent with elevation ordering. Further, the post-processed labels can help in obtaining a better estimate of the elevation ordering, thus resulting in an iterative solution that \emph{simultaneously} infers the elevation ordering and produces physically consistent classification maps. This approach was successfully used in \cite{khandelwal2015post, rse} to build global maps of surface water dynamics. Figure \ref{fig:lake_ex} illustrates the effectiveness of this approach using an example lake in Africa, where the post-processed classification map does not suffer from the errors of the initial classification map and visually matches well with the remote sensing image of the water body.

\begin{figure}[t]
\centering
\includegraphics[width=\linewidth]{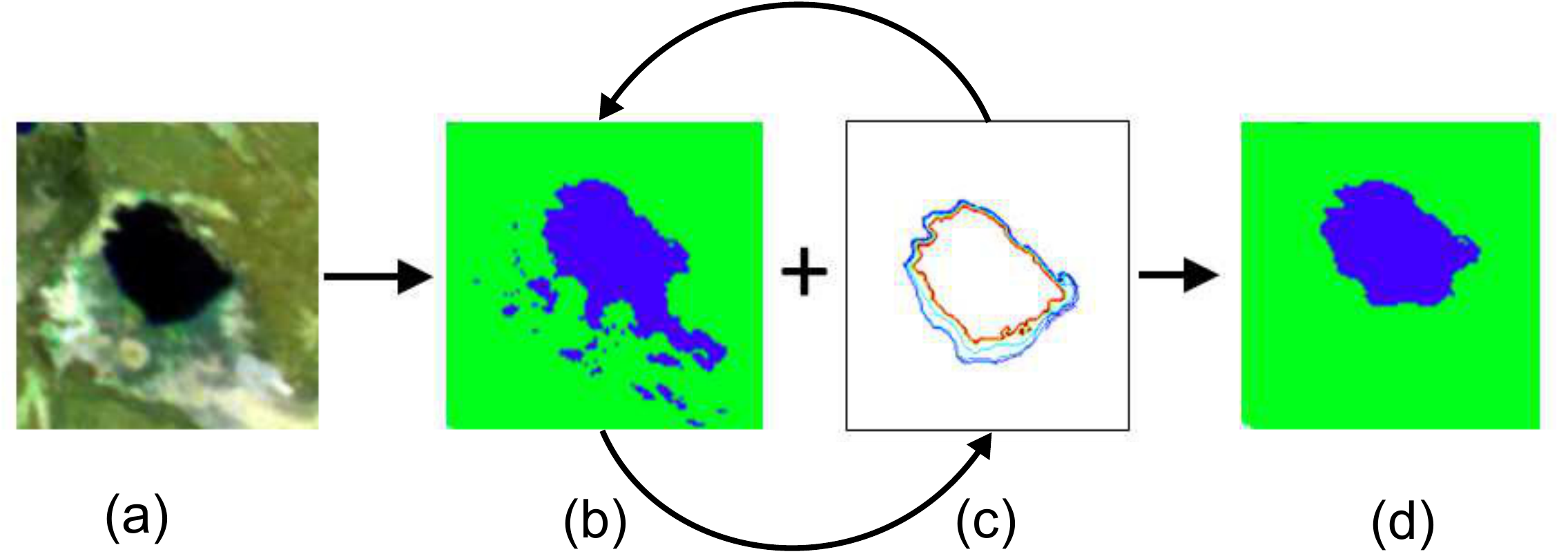}
\caption{Mapping the extent of Lake Abhe (on the border of Ethiopia and Djibouti in Africa) using implicit theory-guided constraints. (a) Remote sensing image of the water body (prepared using multi-spectral false color composites). (b) Initial classification maps. (c) Elevation contours inferred from the history of classification labels. (d) Final classification maps refined using elevation-based constraints.}
\label{fig:lake_ex}
\end{figure}

\IEEEQED
\end{example}

Other examples of the use of implicit constraints includes mapping urbanization \cite{mithal2013change} and tree plantation conversions \cite{xiaowei_tr, xiaowei_sdm}, where hidden Markov models were used to incorporate domain knowledge about the transitions among land covers.

\section{Learning Hybrid Models of Theory and Data Science}
\label{sec:hybrid}


One way to combine the strengths of scientific knowledge and data science is by creating \emph{hybrid} combinations of theory-based and data science models, where some aspects of the problem are handled by theory-based components while the remaining ones are modeled using data science components. There are several ways of fusing theory-based and data science models to create hybrid TGDS models. One way is to build a two-component model where the outputs of the theory-based component are used as inputs in the data science component. This idea is used in climate science for statistical downscaling of climate variables \cite{wilby1998statistical}, where the climate model simulations, available at coarse spatial and temporal resolutions, are used as inputs in a statistical model to predict the climate variables at finer resolutions. Theory-based model outputs can also be used to supervise the training of data science models, by providing physically consistent estimates of the target variable for every training instance.


An alternate way of creating a hybrid TGDS model is to use data science methods to predict intermediate quantities in theory-based models that are currently being missed or inaccurately estimated.
By feeding data science outputs into theory-based models, such a hybrid model can not only show better predictive performance but also amend the deficiencies in existing theory-based models. Further, the outputs of theory-based models may also be used as training samples in data science components \cite{sadowski2016synergies}, thus creating a two-way synergy between them.
Depending on the nature of the model and the requirements of the application, there can be multiple ways of introducing data science outputs in theory-based models. In the following, we provide an illustrative example of this theme of TGDS research in the field of turbulence modeling.



\begin{example}[Turbulence Modeling]
\label{ex:turbulence}
One of the important problems in aerospace engineering is to model the characteristics of turbulent flow, which consists of chaotic changes in the flow velocity, and complex dissipation of momentum and energy. Turbulence modeling is used in a number of applications such as the design and reliability assessment of airfoils in aeroplanes and space vehicles. Key to the study of fluid dynamics is the Navier--Stokes equations, which describe the behavior of viscous fluids under motion. Although the Navier--Stokes equations can be readily applied in simple flow problems involving incompressible and irrotational flow, obtaining an exact representation for turbulent flow requires computationally expensive solutions such as direct numerical simulations (DNS) at fine spatial grids. The high computational costs of DNS make it infeasible for studying practical turbulence problems in the industry, which are typically solved using inexact but computationally cheap approximations. One such approximation is the Reynolds--averaged Navier--Stokes (RANS) equations, which introduces a term called as the Reynolds stress, $\tau$, to represent the apparent stress due to fluctuations caused by turbulence.
Since the exact form of the Reynolds stress is unknown, different approximations of $\tau$ have been explored in previous studies, resulting  in a variety of RANS models. Despite the continued efforts in approximating $\tau$, current RANS models are still insufficient for modeling complex flows with separation, curvature, or swirling. To overcome their limitations, recent work by Wang et al. \cite{wang2016physics} explored the use of machine learning methods to assist RANS models and reduce their discrepancies. In particular, the Reynolds stress was approximated as
\begin{equation}
\tau = \tau_{RANS} + \Delta \tau_{ML},
\end{equation}
where $\tau_{RANS}$ is obtained from a RANS model while $\Delta \tau_{ML}$ is the model discrepancy that is estimated using a random forest model. Although this approach can be used with any generic RANS model to estimate its discrepancy, it does not alter the form of approximation used in obtaining $\tau_{RANS}$, since $\Delta \tau_{ML}$ is learned independently of $\tau_{RANS}$. In another work by Singh et al. \cite{singh2016machine}, a machine learning component was used to directly augment a RANS approximation  in the following manner:
\begin{eqnarray}
-\tau_{ij} &=& 2\rho \nu S_{ij}^{*}-{\frac {2}{3}}\rho K\delta _{ij}, \label{eqn:Boussinesq} \\
\frac{D\nu}{Dt} &=& \beta \times \mathbf{P} - \mathbf{D} + \mathbf{T}, \label{eqn:sa}
\end{eqnarray}
where Equation \ref{eqn:Boussinesq} is the standard Boussinesq equation relating the Reynolds stress $\tau_{ij}$ to the effective viscosity $\nu$, and Equation \ref{eqn:sa} is a variant of the Spalart Allmaras model that estimates $\nu$ as a function of a machine learning term, $\beta$ (learned using an artificial neural network), and other physical terms, $\mathbf{P}$, $\mathbf{D}$, and $\mathbf{T}$, corresponding to production, destruction, and transport processes, respectively. This class of modeling framework, which integrates machine learning terms in theory-based models, has been called field inversion and machine learning (FIML) \cite{parish2016paradigm}.

Both these works illustrate the potential of coupling data science outputs with theory-based models to reduce model discrepancies in complex scientific applications. The exact choice of the data science model and its contribution to the theory-based model can be explored in future investigations. Similar lines of TGDS research can be explored in other domains where current theory-based models are lacking, e.g., hydrological models for studying subsurface flow \cite{ghasemizade2013subsurface}.

\IEEEQED
\end{example}
 
\section{Augmenting Theory-based Models using Data Science}
\label{sec:assimilation}

There are many ways we can use data science methods to improve the effectiveness of theory-based models.
Data can be assimilated in theory-based models for improved selection of model states in numerical models. Data science methods can also help in calibrating the parameters of theory-based models so that they  provide a better realization of the physical system. We describe both these approaches in the following.

\subsection{Data Assimilation in Theory-based Models}

One of the long-standing approaches of the scientific community for integrating data in theory-based models is to use data assimilation approaches, which has been widely used in climate science and hydrology \cite{evensen2009data}. These domains typically involve dynamical systems, such as the progression of climate phenomena over time, which can be represented as a sequence of physical states in numerical models. Data assimilation is a way to infer the most likely sequence of states such that the model outputs are in agreement with the observations available at every time-step. In data assimilation, the values of the current state are constrained to depend on previous state values as well as the current data observations. For example, if we use the Gaussian distribution to model the linear transition between consecutive states, this translates to a Kalman filter. However, in general, the dependencies among the states in data assimilation methods are modeled using more complex forms of distributions that are governed by physical laws and equations. Data assimilation provides a promising step in the direction of integrating data with theory-based models so that the knowledge discovery approach relies both on scientific knowledge and observational data.

\subsection{Calibrating Theory-based Models using Data}
Theory-based models often involve a large number of parameters in their equations that need to be calibrated in order to provide an accurate representation of the physical system. A na\"ive approach for model calibration is to try out every combination of parameter values, perhaps by searching over a discrete grid defined over the parameters,  and choose the combination that produces the maximum likelihood for the data. However, this approach is practically infeasible when the number of parameters are large and every parameter takes many possible values. A number of computationally efficient approaches have been explored in different disciplines for parsimoniously calibrating model parameters with the help of observational data. For example, a seminal work on model calibration in the field of hydrology is the Generalized Likelihood Uncertainty Estimation (GLUE) technique \cite{beven1992future}. This approach models the uncertainty associated with every parameter combination using Monte Carlo approaches, and uses a Bayesian formulation to incrementally update the uncertainties as new observations are made available.
At any given iteration, the parameter combination that shows maximum agreement with the observations is employed in the model, the results of which are used to update the uncertainties on the next iteration. 

The problem of parameter selection has recently received considerable attention in the machine learning community in the context of multi-armed bandit problems \cite{chapelle2011empirical,li2010contextual, jordan2015machine}. The basic objective in these problems is to incrementally select parameter values so that we can \emph{explore} the space of parameter choices and \emph{exploit} the parameter choice that provides the maximum reward, using a limited number of observations.
Variants of these techniques have also been explored for settings where the parameters take continuous values instead of discrete steps \cite{agrawal1995continuum, kleinberg2004nearly}.
These techniques provide a promising direction for calibrating the high-dimensional parameters of theory-based models.

\section{Conclusion}\label{sec:conclusion}


In this paper, we formally conceptualized the paradigm of theory-guided data science (TGDS) that seeks to exploit the promise of data science without ignoring the treasure of knowledge accumulated in scientific principles. We provided a taxonomy of ways in which scientific knowledge and data science can be brought together in any application with some availability of domain knowledge. These approaches range from methods that strictly enforce physical consistency in data science models (e.g., while designing model architecture or specifying theory-based constraints) to methods that allow a relaxed usage of scientific knowledge where our scientific understanding is weak (e.g., as priors or regularization terms). We presented examples from diverse disciplines to illustrate the various research themes of TGDS and also discussed several avenues of novel research in this rapidly emerging field.

One of the central motivations behind TGDS is to ensure better generalizability of models (even when the problem is complex and data samples are under-representative) by anchoring data science algorithms with scientific knowledge.
TGDS also aims at advancing our knowledge of the physical world by producing scientifically interpretable models. Reducing the search space of the learning algorithm to physically consistent models may also have an additional benefit of reducing the computational cost of the algorithm.

The TGDS research themes are not exhaustive and we anticipate the development of novel TGDS themes in the future that explore innovative ways of blending scientific theory with data science.
While most of the discussion in this paper focuses on supervised learning problems, similar TGDS research themes can be explored for other traditional tasks of data mining, machine learning, and statistics. For example, the use of physical principles to constrain spatio-temporal pattern mining algorithms has been explored in \cite{faghmous2012cidu, faghmous2014aaai} for finding ocean eddies from satellite data.
The need to explore TGDS models for uncertainty quantification is discussed in \cite{npg-21-777-2014} in the context of understanding and projecting climate extremes. 
Scientific knowledge can also be used to advance other aspects of data science, e.g., the design of scientific work-flows \cite{honavar2014promise, gil2007examining} or the generation of model simulations \cite{paganini2017calogan}.

We hope that this paper serves as a first step in building the foundations of TGDS and encourages follow-on work to develop in-depth theoretical formalizations of this paradigm. While success in this endeavor will need significant innovations in our ability to handle the diversity of forms in which scientific knowledge is represented and ingested in different disciplines (e.g., differences in granularity and type of information, degree of completeness, and uncertainty in knowledge), the concrete TGDS approaches presented in this paper can be considered as a stepping stone in this ambitious journey. We anticipate the deep integration of theory-based and data science to become a quintessential tool for scientific discovery in future research. The paradigm of TGDS, if effectively utilized, can help us realize the vision of the ``fourth paradigm'' \cite{hey2009fourth} in its full glory, where data serves an integral role at every step of scientific knowledge discovery.
\ifCLASSOPTIONcaptionsoff
  \newpage
\fi

\ifCLASSOPTIONcompsoc
  \section*{Acknowledgments}
\else
  \section*{Acknowledgment}
\fi

The ideas in this vision paper were developed while being funded by an NSF Expeditions in Computing Grant \#1029711.

\bibliographystyle{IEEEtran}
\bibliography{IEEEabrv,bibs/ieee_tkde,bibs/new,bibs/anuj,bibs/anuj2}


\begin{IEEEbiography}[{\includegraphics[height=1in]{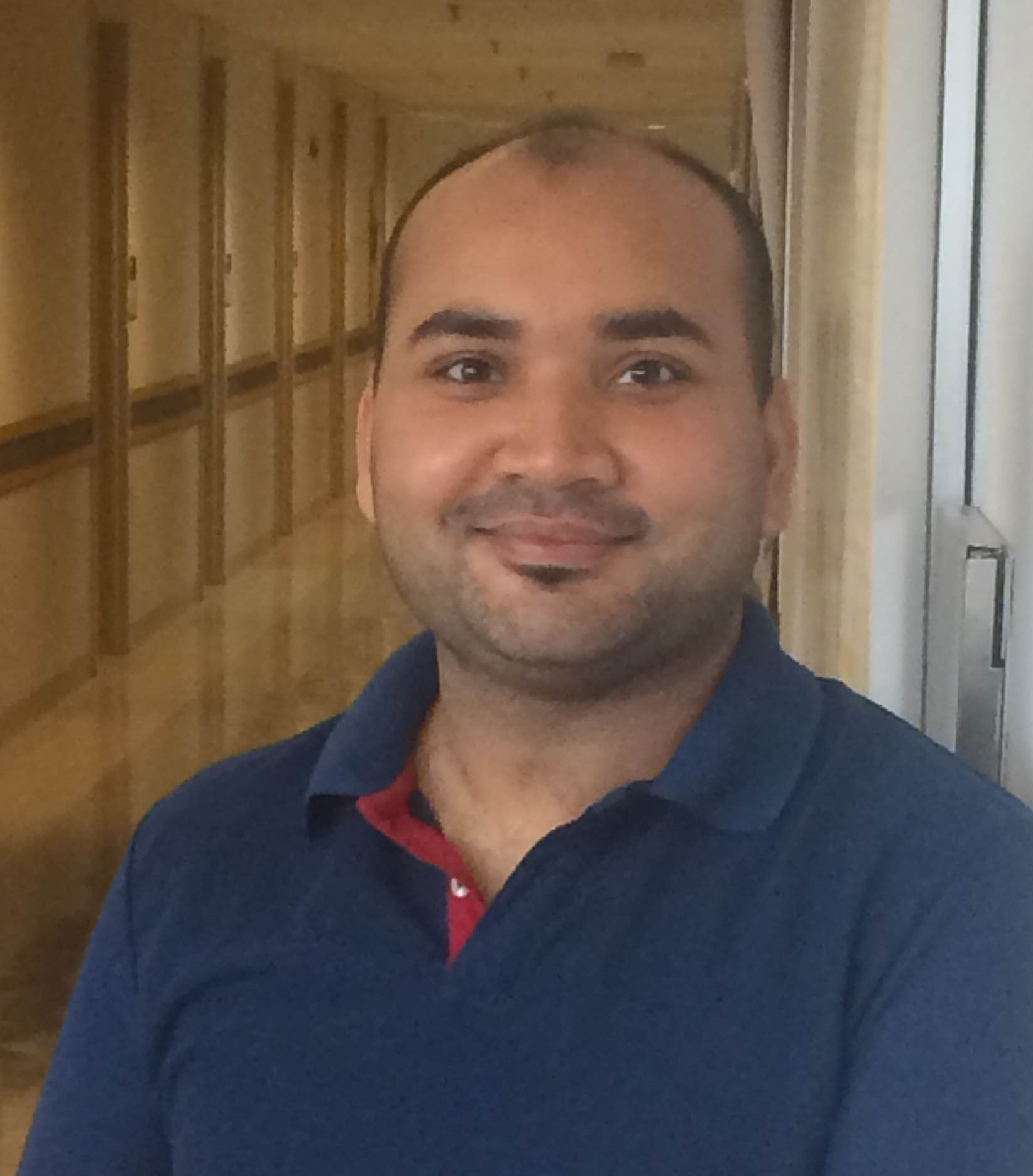}}]{Anuj Karpatne}
Anuj Karpatne is a PhD candidate in the Department of Computer Science and Engineering (CSE) at University of Minnesota (UMN). 
Karpatne works in the area of spatio-temporal data mining for enviornmental applications. 
Karpatne received his B.Tech-M.Tech degree in Mathematics \& Computing from Indian Institute of Technology (IIT) Delhi.
\end{IEEEbiography}

\vspace{-12ex}

\begin{IEEEbiography}[{\includegraphics[height=1in]{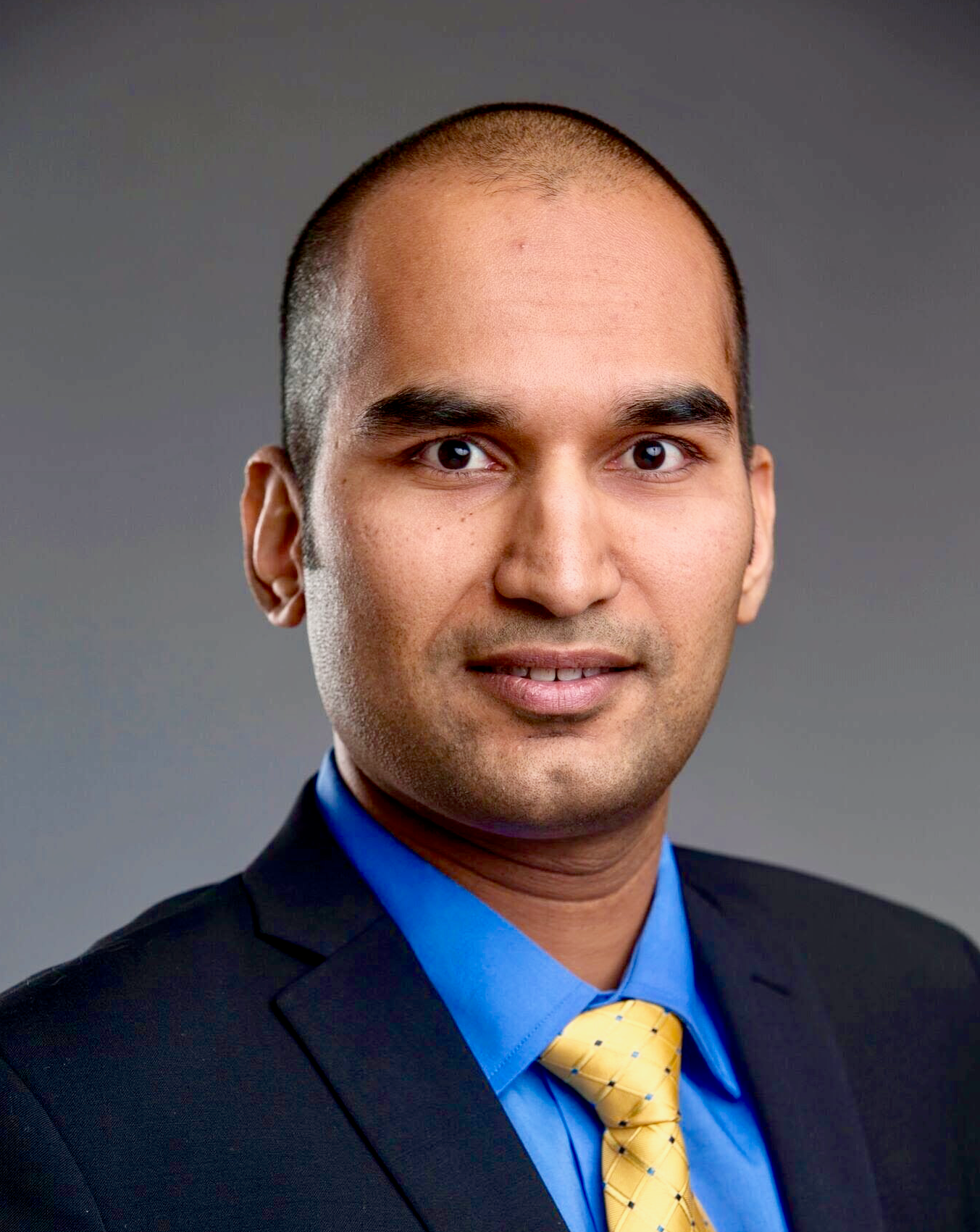}}]{Gowtham Atluri}
Gowtham Atluri is an Assistant Professor in the Department of Electrical Engineering and Computer Science (EECS) at University of Cincinnati. 
Atluri's research interests include data mining, neuroimaging, and climate science. 
Atluri received his PhD in Computer Science (CS) from UMN and M.Tech in CS from IIT Roorkee.
\end{IEEEbiography}

\vspace{-12ex}

\begin{IEEEbiography}[{\includegraphics[height=0.9in]{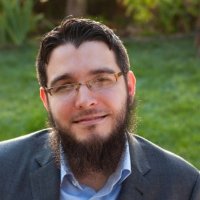}}]{James Faghmous}
James Faghmous is an Assistant Professor and Founding CTO of the Arnhold Global Health Institute at 
Icahn School of Medicine at Mount Sinai.
Faghmous' research interests include data science, climate science, and global health. 
Faghmous received his PhD in CS and M.S in CS from UMN, and B.S in CS from City College of New York.
\end{IEEEbiography}

\vspace{-10ex}

\begin{IEEEbiography}[{\includegraphics[height=0.8in]{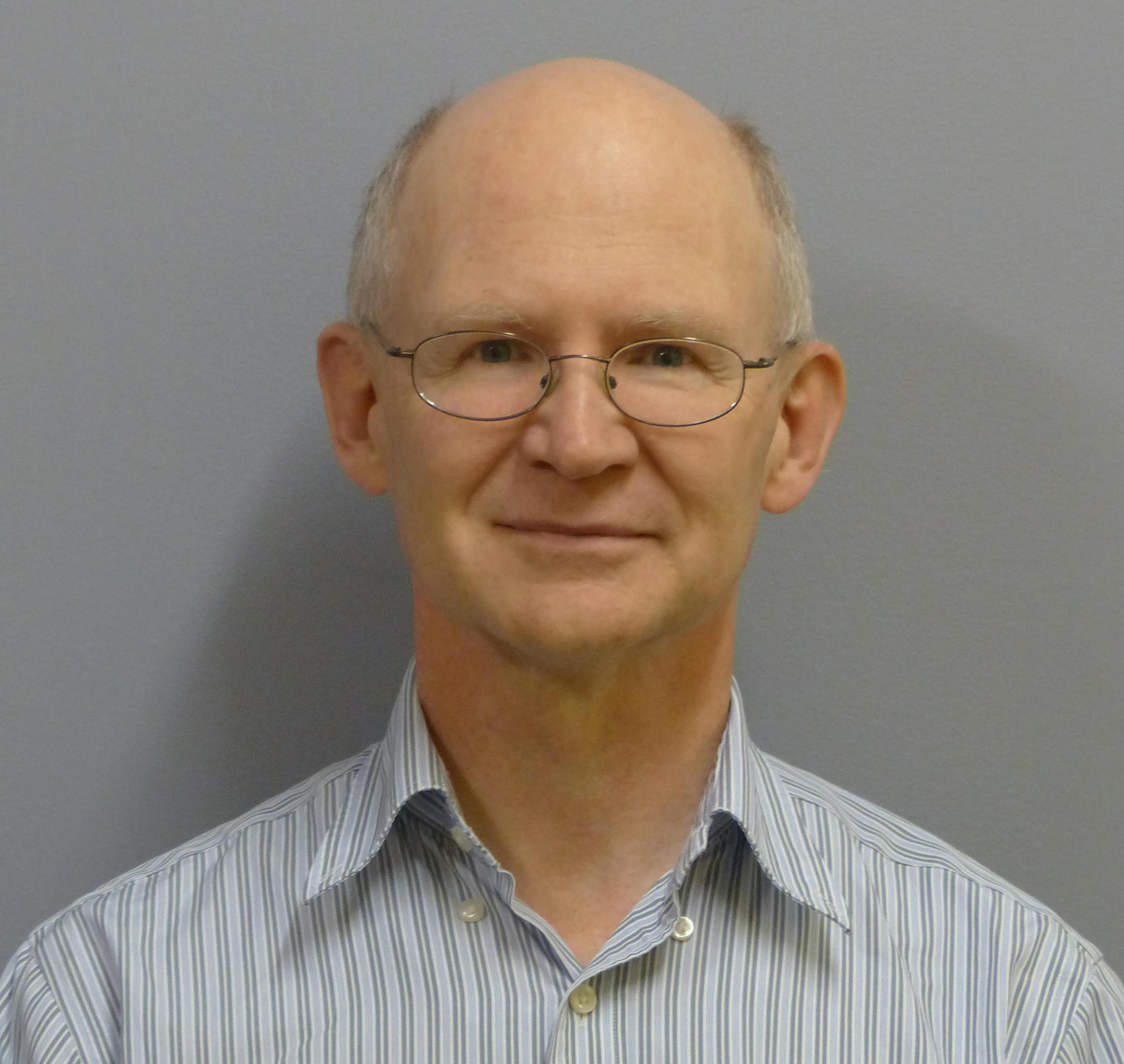}}]{Michael Steinbach}
Michael Steinbach is a Research Associate in the Department of CSE at UMN.
Steinbach's research interests include data mining, healthcare, bio-informatics, and statistics. 
Steinbach received his PhD in CS, M.S in CS and Statistics, and B.S in Math from UMN.
\end{IEEEbiography}

\vspace{-12ex}

\begin{IEEEbiography}[{\includegraphics[width=0.9in]{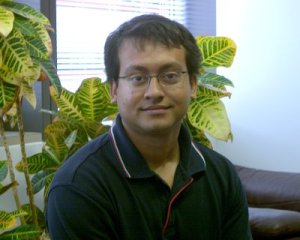}}]{Arindam Banerjee}
Arindam Banerjee is an Associate Professor in the Department of CSE at UMN.
Banerjee's research interests include machine learning, data mining, and optimization.
Banerjee received his PhD in CS from University of Texas Austin, M.Tech in Electrical Engineering (EE) from IIT Kanpur (IITK), and B.E in EE from Jadavpur University.
\end{IEEEbiography}

\vspace{-12ex}

\begin{IEEEbiography}[{\includegraphics[height=1in]{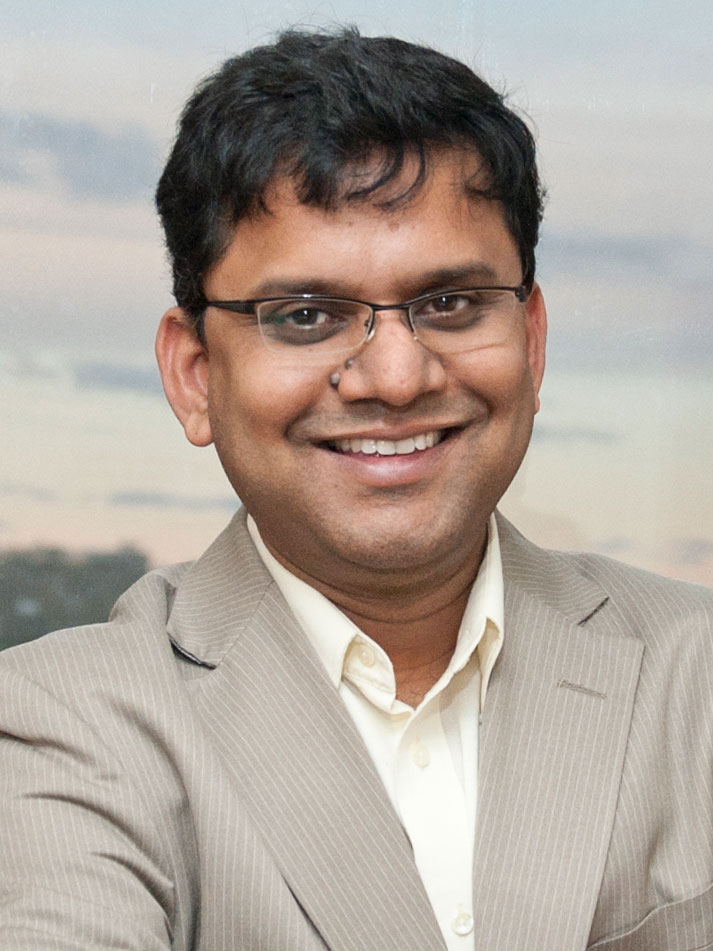}}]{Auroop Ganguly}
Auroop Ganguly is an Associate Professor in the Department of Civil and Environmental Engineering (CEE) at Northeastern University. Ganguly's research encompasses weather extremes, water sustainability, and the resilience of critical infrastructures.
Ganguly received his PhD in CEE from Massachusetts Institute of Technology.
\end{IEEEbiography}

\vspace{-12ex}

\begin{IEEEbiography}[{\includegraphics[height=1in]{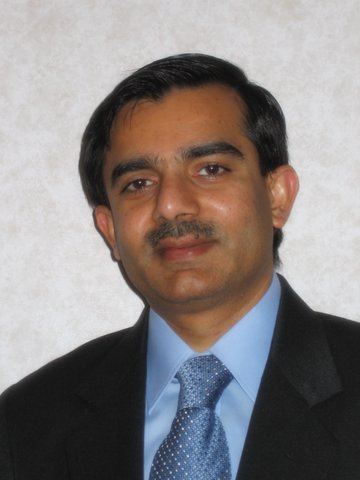}}]{Shashi Shekhar}
Shashi Shekhar is a McKnight Distinguished University Professor in the Department of CSE at UMN.
Shekhar's research interests include spatial databases, spatial data mining, and geographic and information systems.
Shekhar received his PhD in CS and M.S in Business Administration from University of California Berkeley, and B.Tech in CS from IITK.
\end{IEEEbiography}

\vspace{-12ex}

\begin{IEEEbiography}[{\includegraphics[height=0.9in]{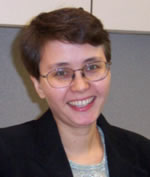}}]{Nagiza Samatova}
Nagiza Samatova is a Professor in the Department of CS at North Carolina State University.
Samatova's research interests include theory of computation, data science, and high performance computing.
Samatova received her PhD in Applied Mathematics from Computational Center of Russian Academy of Sciences Moscow. 
\end{IEEEbiography}

\vspace{-12ex}

\begin{IEEEbiography}[{\includegraphics[height=1in]{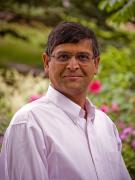}}]{Vipin Kumar}
Vipin Kumar is a Regents Professor and William Norris Chair in Large Scale Computing in the Department of CSE at UMN.
Kumar's research interests include data mining, high-performance computing, and their applications in climate/ecosystems and biomedical domains.
Kumar received his PhD in CS from University of Maryland, M.E in EE from Philips International Institute Eindhoven, and B.E in Electronics \& Communication Engineering from IIT Roorkee.
\end{IEEEbiography}
\end{document}